\documentclass{article}

\usepackage{microtype}
\usepackage{graphicx}
\usepackage{subfigure}
\usepackage{booktabs} 

\usepackage{multirow}
\usepackage{longtable}
\usepackage{amsmath}
\usepackage{tabularx}

\usepackage{hyperref}

\usepackage[accepted]{icml2021}

\icmltitlerunning{BANG: Bridging Autoregressive and Non-autoregressive Generation with Large Scale Pretraining}

\begin{document}

\twocolumn[
\icmltitle{BANG: Bridging Autoregressive and Non-autoregressive Generation with Large Scale Pretraining}



\icmlsetsymbol{equal}{*}

\begin{icmlauthorlist}
\icmlauthor{Weizhen Qi}{ustc,intern}
\icmlauthor{Yeyun Gong}{equal,msra}
\icmlauthor{Jian Jiao}{equal,microsoft}
\icmlauthor{Yu Yan}{equal,microsoft}
\icmlauthor{Weizhu Chen}{equal,microsoft}
\icmlauthor{Dayiheng Liu}{equal,intern,scu}
\icmlauthor{Kewen Tang}{microsoft}
\icmlauthor{Houqiang Li}{ustc}
\icmlauthor{Jiusheng Chen}{microsoft}
\icmlauthor{Ruofei Zhang}{microsoft}
\icmlauthor{Ming Zhou}{msra}
\icmlauthor{Nan Duan}{msra}
\end{icmlauthorlist}

\icmlaffiliation{ustc}{University of Science and Technology of China, Hefei, China} 
\icmlaffiliation{intern}{During Internship at MSRA}  
\icmlaffiliation{msra}{Microsoft Research Asia, Beijing, China}  
\icmlaffiliation{microsoft}{Microsoft, Redmond, USA}  
\icmlaffiliation{scu}{Sichuan University, Chengdu, China}

\icmlcorrespondingauthor{Yeyun Gong}{yegong@microsoft.com}

\icmlkeywords{Natural Language Generation, Pretraining}

\vskip 0.3in
]



\printAffiliationsAndNotice{\icmlEqualContribution} 

\begin{abstract}

In this paper, we propose \textbf{BANG}, a new pretraining model to \textbf{B}ridge the gap between \textbf{A}utoregressive (AR) and \textbf{N}on-autoregressive (NAR) \textbf{G}eneration. AR and NAR generation can be uniformly regarded as to what extent previous tokens can be attended, and BANG bridges AR and NAR generation by designing a novel model structure for large-scale pretraining. The pretrained BANG model can simultaneously support AR, NAR and semi-NAR generation to meet different requirements.  Experiments on question generation (SQuAD 1.1), summarization (XSum) and dialogue generation (PersonaChat) show that BANG improves NAR and semi-NAR performance significantly as well as attaining comparable performance with strong AR pretrained models. Compared with the semi-NAR strong baselines, BANG achieves absolute improvements of 14.01 and 5.24 in the overall scores of SQuAD 1.1 and XSum, respectively. In addition, BANG achieves absolute improvements of 10.73, 6.39 and 5.90 in the overall scores of SQuAD, XSUM and PersonaChat respectively compared with the strong NAR baselines.

\end{abstract}

\section{Introduction}\label{sec.introduction}
Various pretraining methods~\cite{song2019mass,lewis2019bart,yan2020prophetnet,raffel2020exploring,zhang2019pegasus} have been successfully applied in natural language generation. Most of the pretraining works are based on Transformer and designed with autoregressive (AR) language model. Transformer based pretraining models show consistent improvements with larger model size and larger pretraining corpus. Although the autoregressive generation method achieves high-quality results in many tasks, its latency is a well-known limitation for online real-time usage. 

Non-autoregressive (NAR) models~\cite{gu2017non,lee2018deterministic,ghazvininejad2019mask,raffel2020exploring,zhang2019pegasus} are proposed to reduce generation latency. Different from AR models which generate tokens sequentially, NAR models generate tokens in parallel. Compared to AR models, NAR models generally come with a much lower inference latency, but a decrease in accuracy. In order to balance latency and accuracy, semi-NAR generation models~\cite{stern2019insertion, lee2018deterministic, gu2019levenshtein,ghazvininejad2019mask} are proposed. However, most of the NAR and semi-NAR models focus on translation tasks rather than general natural langauge generation tasks, which are proved to significantly benefit from pretraining~\cite{yan2020prophetnet, lewis2019bart}. Some works~\cite{guo2020incorporating, su2021non} initialize their NAR models with pretrained natural language understanding model BERT~\cite{devlin2018bert} for better performance. To the best of our knowledge, this paper proposes the first large-scale pretraining model designed for NAR and semi-NAR generation.

In this paper, we propose a new model named BANG \footnote{https://github.com/microsoft/BANG} to bridge the gap between AR and NAR via pretraining a generative model. Specifically, we consider pretraining model using AR, semi-NAR and NAR objectives with different attention mechanisms, which decide what extent previous tokens can be attended to. Precisely, BANG is pretrained to predict each token with arbitrary length of previous golden tokens replaced with special token [MASK]s. For example, with complete previous golden tokens, BANG predicts the next token in the AR manner. With all previous tokens replaced by [MASK], BANG predicts the next token in the NAR manner. 

For AR models, the training strategy of teacher-forcing is commonly used, which uses the golden tokens as previous context to predict the next token. For NAR models, [MASK] initialization~\cite{ghazvininejad2019mask} or other methods like encoder copy~\cite{gu2017non} and posterior distribution approximation~\cite{shu2020latent} are applied.  In BANG pretraining, we consider the previous context of golden and [MASK] tokens, with arbitrary golden tokens' length and [MASK]  tokens' length. To achieve an efficient implementation for multiple arbitrary alternatives in a same output sequence, we propose a new structure named cross-stream visible n-stream self-attention, which can be used to train BANG with different golden and [MASK] tokens' combinations. For usage on downstream tasks, the single pretrained BANG model can be directly finetuned for either vanilla AR models or vanilla NAR models. Additionally, BANG can also be finetuned for hybrid semi-NAR models, which support predicting tokens with arbitrary previous golden tokens or [MASK]. Concretely, for semi-NAR generation, BANG predicts the first several tokens one by one as a high-quality sub-sequence hint, then produces all the remaining tokens simultaneously.

Our main contributions are: 
1) BANG bridges the gap between AR and NAR by considering arbitrary previous [MASK] length during large-scale pretraining. 2) BANG is pretrained using an efficient cross-stream visible n-stream decoder to realize parallelization. Given multiple arbitrary number of previous tokens replaced with [MASK], every token is trained to predict simultaneously at each time step. 
3) BANG supports NAR, semi-NAR and AR finetuning to meet different requirements with the same pretrained model structure. 
4) We pretrain BANG with 16GB English language corpora of Wikipedia and BookCorpus, and finetune it on 3 popular natural language generation tasks in AR, semi-NAR and NAR manners, respectively. For NAR and semi-NAR finetuning, BANG achieves significant performance improvements on all the tasks. For AR finetuning with the comparison to strong AR pretrained models, BANG can attain comparable performance. 

\section{Model Structure}\label{sec.model.structure}

In this section, we first introduce fundamental concepts in~\S~\ref{sec.fundamental.concepts}, and then describe BANG model structure and implementation in~\S~\ref{sec.bang}. Finally the finetuning strategies are given in ~\S~\ref{sec.finetune}.

\subsection{Preliminaries}\label{sec.fundamental.concepts}

Natural language generation typically refers to the procedure of generating an output sequence $Y=\{y_1, y_2, ..., y_{|Y|}\}$ given an input sequence $X=\{x_1, x_2, ..., x_{|X|}\}$. There are three common generative paradigms: AR, NAR and semi-NAR generation. 

\paragraph{AR generation} follows the conditional probability:
\begin{equation}
P(Y|X) =  \prod_{t=1}^{|Y|} P(y_t|y_{< t},X)  
\label{prob.s2s.ar} 
\end{equation}

Each token $y_t$ in the target sequence $Y$ is predicted with the dependency of input sequence $X$ and previous tokens $y_{< t}$. Vanilla Transformer realizes this target by shifting decoder inputs one position, each token attending to its previous tokens to predict the next token.

\paragraph{NAR generation} follows the conditional probability:
\begin{equation}
P(Y|X) =  \prod_{t=1}^{|Y|} P(y_t|X)  
\label{prob.s2s.nar} 
\end{equation}

Each token $y_t$ in target sequence $Y$ is predicted with the dependency of input sequence $X$ but no dependency of previous tokens. Each token in the target sequence should be predicted simultaneously. An intuitive strategy is to feed a list of [MASK] tokens as the decoder initialization \cite{ghazvininejad2019mask}.

\paragraph{Semi-NAR generation} follows the conditional probability:
\begin{equation}
P(Y|X) =  \prod_{t=1}^{|Y|} P(y_t|Y_{c_t}, X)  
\label{prob.s2s.seminar} 
\end{equation}

Here, $Y_{c_t}$ means visible context for $y_t$ from target sequence $Y$. $Y_{c_t}$ is designed differently in different semi-NAR models~\cite{stern2019insertion, lee2018deterministic, ghazvininejad2019mask, gu2019levenshtein}.

\subsection{BANG}\label{sec.bang}

\begin{figure*}[ht]
    \centering
    \includegraphics[width = 6.0in]{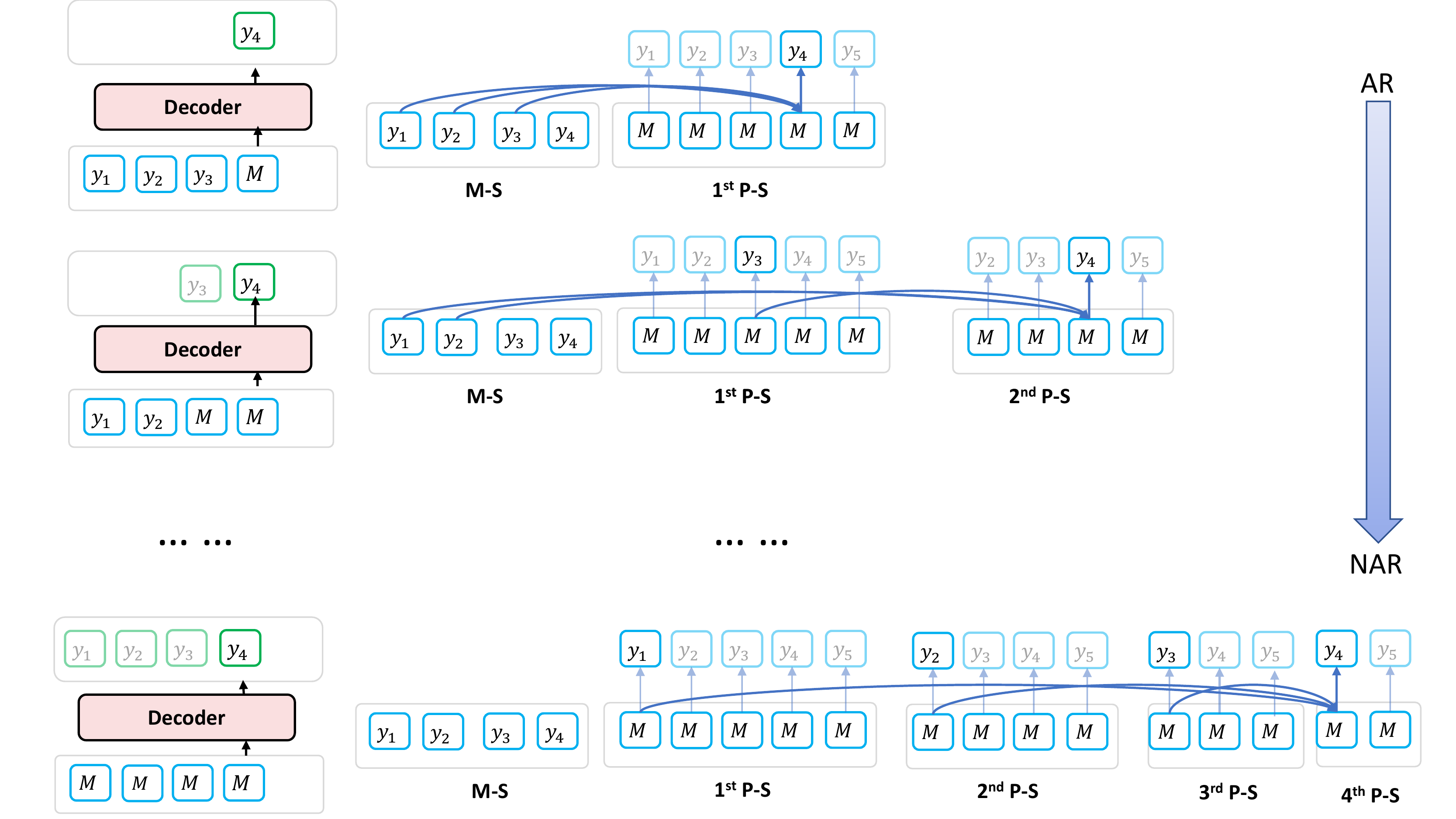}
	\caption{BANG pretraining. The prediction of $y_4$ is used as an example. ``M-S'' and ``P-S'' represent ``Main stream" and ``Predicting stream'', respectively. The $1^{st}$ predicting stream predicts tokens in AR manner. First tokens in each predicting stream compose NAR prediction.}
	\label{bang.pretrain}
\end{figure*}

In this section, we will introduce the overall model structure of BANG in~\S~\ref{sec.overall} and the proposed cross-stream visible n-stream self-attention in~\S~\ref{sec.cross.visible}. 

\subsubsection{Overall}\label{sec.overall}
BANG is a sequence to sequence model based on the Transformer Encoder-Decoder structure. It is composed of a multi-layer Transformer encoder with vanilla self-attention mechanism  and a multi-layer Transformer decoder with proposed cross-stream visible n-stream self-attention mechanism. We consider the input sequence $X=\{x_1, x_2, \dots , x_{|X|}\}$ and output sequence $Y=\{y_1, y_2, \dots, y_{|Y|}\}$. On the encoder side, BANG firstly encodes $X$ into a list of hidden states by stacked vanilla Transformer encoder layers:
\begin{equation}
H_{\rm{enc}} = \textbf{Encoder}(x_1, \dots, x_{|X|}), 
\label{eq:encoder} 
\end{equation}
On the decoder side, BANG predicts each token $y_t$ in output sequence $Y$ with cross-stream visible n-stream self-attention. Besides predicting each token with all previous golden tokens in the AR manner and all [MASK]s in a NAR manner, BANG targets at predicting them with arbitrary length of previous golden tokens replaced with [MASK]s as the context. Formally, for $y_t$ in $Y$:
\begin{equation}
\begin{aligned}
P(y_t|y_{<t},x),  P(y_t|y_{<t-1},x),  P(y_t|y_{<t-2},x),    \\
\dots,  P(y_t|y_1, x),  P(y_t|x) = \textbf{Decoder}(y_{<t},H_{\rm{enc}}) 
\end{aligned}
\label{eq.decoder} 
\end{equation}
The conditional probability of target sequence in BANG can be described as:
\begin{equation}
P(\hat{Y}|X) = \prod_{t=1}^{|Y|}{  \prod_{i=1}^{t} P(y_t|y_{\leq t-i},X)  }
\label{prob.s2s.lm1} 
\end{equation}
BANG targets at optimizing $\hat{Y}$ rather than the original output sequence $Y$. For each token $y_t$ in $Y$, $\hat{Y}$ considers replacing its $i$ previous tokens with [MASK] for any $i<t$. For maximum likelihood training with a cross-entropy loss:

\begin{equation}
\begin{aligned}
\mathcal{L}_{\textrm{BANG}} & = logP(\hat{Y}|X)  \\
& = \sum_{t=1}^{|Y|}{  \sum_{i=1}^{t} logP(y_t|y_{\leq t-i},X)  } \\
& = \underbrace{\sum_{t=1}^{|Y|} logP(y_t|y_{<t},X)}_{\text{AR part}} \\
& + \underbrace{\sum_{t=2}^{|Y|}  { \sum_{i=2}^{t-1} logP(y_t|y_{\leq t-i},X)  }}_{\text{Bridging part}} \\
& + \underbrace{\sum_{t=1}^{|Y|} logP(y_t|X)}_{\text{NAR part}} \\
\end{aligned}
\label{prob.s2s.lm2} 
\end{equation}

We can see the BANG optimizing target is composed of three parts: AR part, NAR part and bridging part. AR part and NAR part directly benefit down-stream generation pattern. The bridging part composes a curriculum learning path to benefit NAR from high-accuracy AR learning.

\subsubsection{Cross-stream Visible N-stream Self-attention}\label{sec.cross.visible}
To implement the decoder of BANG, we design a cross-stream visible n-stream self-attention based on the Transformer decoder. There are self-attention sub-layers, encoder-decoder attention sub-layers and feed-forward sub-layers in Transformer decoder layers. Encoder-decoder attention sub-layers and feed-forward sub-layers are used to fetch information from encoded hidden states and do linear calculation. We keep these two kinds of sub-layers unchanged. Self-attention sub-layers control the information flow and we propose cross-stream visible n-stream self-attention to replace it.

We duplicate decoder into one main stream and $n$ predicting streams where $n$ equals to target sequence length by sharing parameters. Main stream is fed with golden tokens as golden previous context. Predicting streams are fed with [MASK]s to predict corresponding tokens and these [MASK]s also serve as previous tokens for NAR generation. 
For the tokens in the $i^{th}$ predicting stream, their $i-1$ previous tokens are replaced by [MASK] from its previous predicting streams, and further previous tokens are golden tokens from the main stream. Each token in the $1^{st}$ predicting stream is predicted in AR pattern with all previous golden tokens from main stream visible. The first tokens in each predicting stream compose the NAR generation with all previous visible tokens as previous predicting streams' [MASK]s. 

Figure~\ref{bang.pretrain} illustrates how BANG bridges the AR-NAR generation with the prediction of $y_4$ as an example. In the $1^{st}$ predicting stream, [MASK] for $y_4$ attends to real $y_1$, $y_2$ and $y_3$ from main stream golden tokens. Each token in the $1^{st}$ predicting stream works the same as $y_4$ and is pretrained in AR manner in parallel. In the $2^{nd}$ predicting stream, [MASK] for $y_4$ attends to real $y_1$ and $y_2$ from main stream golden token and [MASK] for $y_3$ from $1^{st}$ predicting stream. $y_3$ in the $1^{st}$ predicting stream and $y_4$ in the $2^{nd}$ predicting stream are generated with the conditional probability of $P(y_3, y_4 | y_0, y_1)$ as shown on the left of the figure. We can see increasing difficulty in the $2^{nd}$ predicting stream and the transition from AR to NAR as more previous tokens are replaced with [MASK]s. In the $4^{th}$ predicting stream, $y_4$ is predicted with no previous golden tokens visible but only [MASK]s from its previous predicting streams are visible. Note that for each token in the target sequence, arbitrary previous golden tokens replaced with [MASK]s are considered in BANG, and all predicting streams are computed simultaneously. At each time step, all tokens in the target sequence from AR pattern to NAR pattern are predicted simultaneously.

To reduce GPU memory and computational cost, we adopt block-wise attention calculation. Since only the previous predicting streams are visible, each predicting stream is only concatenated with previous streams to avoid extra computation cost. 
The workflow in decoder layer $l$ can be described in Algorithm~\ref{alg.bang}.

\begin{algorithm}[h]
   \caption{Cross-stream Visible N-stream Self-attention}
   \label{alg:example}
\begin{algorithmic}
   \STATE {\bfseries Input:} $h_l$
   \STATE {\bfseries Output:} $h_{l+1}$
   \STATE $K_{\textrm{cache}} \leftarrow \emptyset$ \;
   \STATE $V_{\textrm{cache}} \leftarrow \emptyset$ \;
   \STATE $h_{l+1} \leftarrow \emptyset$ \;
   \STATE $Q, K, V =  Linear(h_l)$ \;
   \FOR{$Q_i, K_i, V_i$ in  split(Q, K, V)  } 
   \STATE $K_{\textrm{cache}} \leftarrow K_{\textrm{cache}} \oplus K_i$ \;
   \STATE $V_{\textrm{cache}} \leftarrow V_{\textrm{cache}} \oplus  V_i$ \;
   \STATE $h_{l+1} \leftarrow h_{l+1} \oplus  Attn(Q_i, K_{\textrm{cache}}, V_{\textrm{cache}})$ \;
   \ENDFOR
   \STATE return $h_{l+1}$ \;
\end{algorithmic}
\label{alg.bang}
\end{algorithm}

Function $Linear$ has three outputs: Q, K, V for self-attention from input hidden states $h_l$. $\oplus$ denotes the concatenate operator, and function $Attn$ is defined in equation~\ref{prob.s2s.atten}:
\begin{equation}
\begin{aligned}
Attn(Q, K, V) = softmax(\frac{QK^T}{\sqrt{d}}+L)V
\end{aligned}
\label{prob.s2s.atten} 
\end{equation}
, where $d$ represents the dimension for $Q$,$K$ and $V$ vectors, $L$ represents the relative positional bias and mask matrix to make sure only visible tokens's $V$ can be attended to.

\subsection{BANG Finetuning}\label{sec.finetune}
In \S~\ref{sec.overall} and \S~\ref{sec.cross.visible}, we introduce the model structure and its language model for pretraining. In this section, we will introduce BANG finetuning strategy for AR, NAR and semi-NAR generation.
\subsubsection{AR generation}
\begin{figure}[h]
    \centering
    \includegraphics[width = 3.0in]{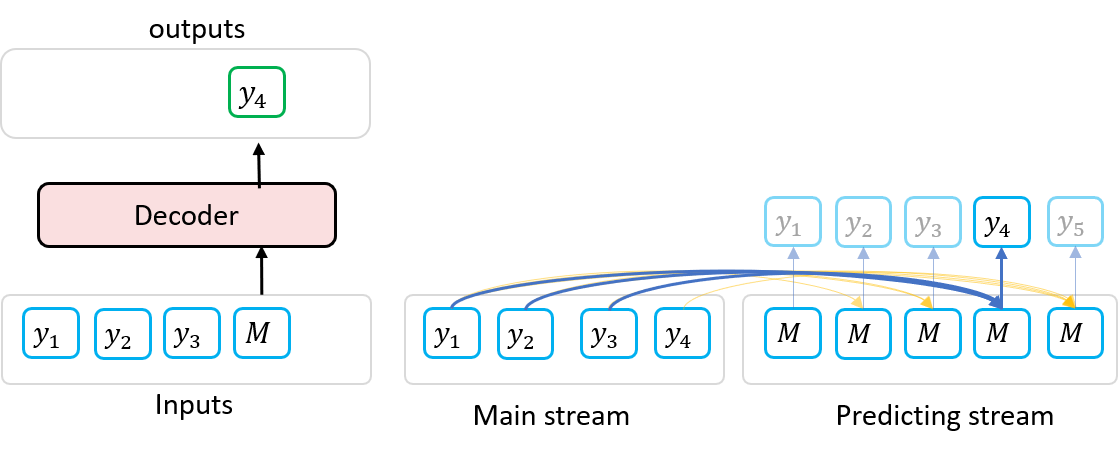}
	\caption{BANG AR generation. The [MASK] can attend to its previous golden tokens to predict $y_4$.} 
	\label{bang.ar}
\end{figure}

BANG AR generation pattern is same as XLNet~\cite{yang2019xlnet} 2-stream strategy. We take the prediction of $y_3$ as an example in Figure~\ref{bang.ar} with the encoder part omitted.  

On the left, BANG predicts $y_4$ with $y_1$, $y_2$ and $y_3$ visible as $P(y_4 | y_1, y_2, y_3)$. During model training, to predict each token in parallel, BANG duplicates decoder layers into one main stream and one predicting stream as shown on the right of Figure~\ref{bang.ar}. [MASK] in the predicting stream attends to its previous tokens from main stream to predict corresponding position's target token. 
\subsubsection{NAR generation}
\begin{figure}[h]
    \centering
    \includegraphics[width = 3.0in]{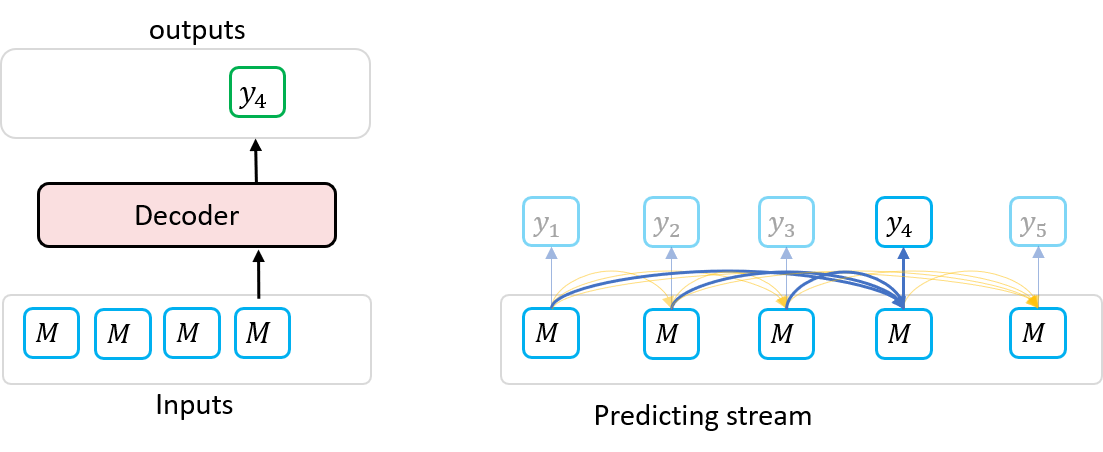}
	\caption{BANG NAR generation. The [MASK] can only attend to its previous [MASK]s to predict $y_3$. }
	\label{bang.nar}
\end{figure}

BANG NAR generation pattern is same as vanilla Transformer. We take the prediction of $y_4$ as an example in Figure~\ref{bang.nar} with the encoder part omitted.

On the left, we can see BANG predicts $y_4$ with no context information from target sequence $Y$ but a list of [MASK]s. On the right, a main difference between BANG NAR and CMLM or NAT is that [MASK] in BANG decoder can only attend to its previous [MASK]s rather than every token in the decoder. We design BANG in this attention manner with two reasons: 1) To be consistent with BANG AR and benefit from AR-NAR bridging pretraining. 2) If [MASK] in the decoder has bi-directional attention, the number of fed [MASK] will influence the result, under which condition a length predictor is needed. If the model predicts the wrong target sequence length, the decoder has to fill all the extra tokens or terminate unexpectedly (early). In BANG, only previous [MASK]s are visible, and the first generated [EOS] token is considered the signal of sentence end token as traditional AR models.

\subsubsection{Semi-NAR generation}

\begin{figure}[h]
    \centering
    \includegraphics[width = 3.0in]{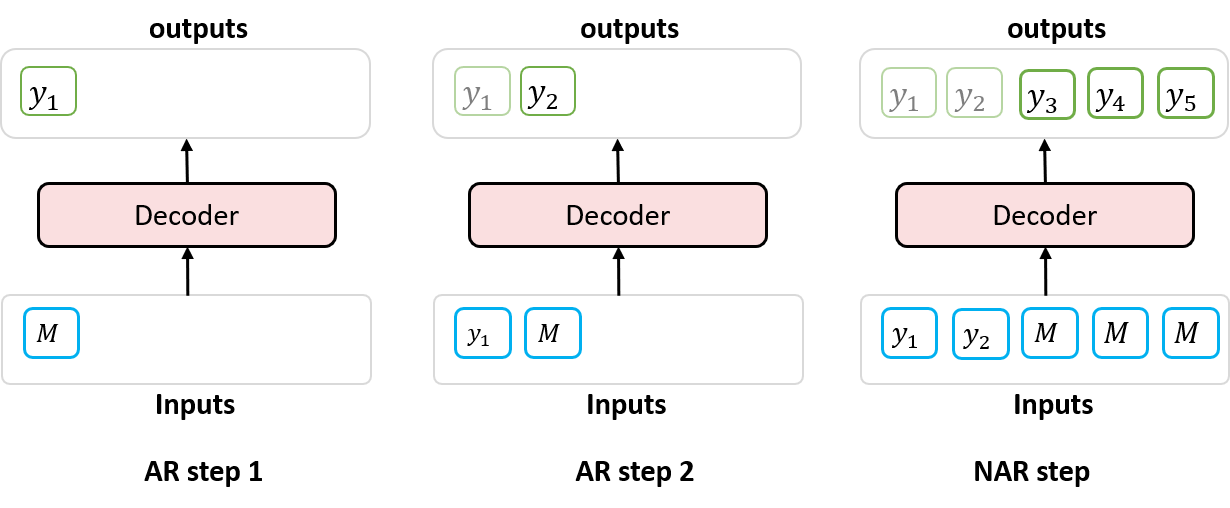}
	\caption{BANG semi-NAR generation. BANG semi-NAR inference supports predicting first several tokens step by step then generating rest tokens at one timestep. Here we present the inference procedure. The training procedure is same to Figure~\ref{bang.pretrain}. } 
	\label{bang.semiar}
\end{figure}

For BANG semi-NAR generation, visible context $Y_{c_t}$ is designed as complete golden context visible for first several tokens in AR manner and golden+[MASK] for the rest tokens. BANG semi-NAR training procedure is same as pretraining implementation introduced in ~\S~\ref{sec.bang} and we show the inference procedure in Figure~\ref{bang.semiar}.

We can see that BANG predicts first two tokens one by one in AR manner. Then the AR generated sub-sequence is used as a hint for the rest tokens. A list of [MASK]s are appended to be predicted at one time step in NAR manner. The motivation of this semi-NAR generation pattern is that NAR models always mix up target sequence different expressions. High-quality AR generated sub-sequence can help NAR step about choosing the writing style and other dependencies. We can also generate the outputs in vanilla AR or NAR manner with this finetuning strategy.

\section{Experiments and Results}
In this section, we first introduce the experimental setup and datasets in~\S~\ref{sec:exp1}, and report the main experimental results in~\S~\ref{sec:exp2}. Next, we conduct a comparison between the NAR pretraining and the BANG pretraining in~\S~\ref{sec.abalation.pretrain}. Last, we put some case studies for each dataset in the Appendix.

\subsection{Setup and Dataset} \label{sec:exp1}

\subsubsection{Setup}

We pretrain BANG using the 16GB corpus (Wikipedia and BookCorpus). The BANG model has a 6-layer encoder and a 6-layer decoder, with a dimension of 768 in their hidden states, to be consistent with AR pretrained models ~\cite{song2019mass, yan2020prophetnet, lewis2019bart} base version model size.

We implement BANG with the span mask and prediction pretraining task, same as MASS~\cite{song2019mass}. In each block of an input sentence, we mask up to $n$ continuous tokens for prediction, where $n$ is same as the number of predicting streams. We mask 15\% tokens for every 64 tokens in the input sequence span, which makes the target sequence length in each span as 9. BANG has 9 predicting streams, same as sequence length. We pretrain BANG from scratch with a learning rate of 3e-4 for 35 epochs and a batch size of 2048. BANG uses the same dictionary as BERT-uncased~\cite{devlin2018bert}.

For AR generation, we load the BANG pretrained model and finetune it with teacher forcing strategy. We use the AR model~\textbf{Transformer}~\cite{vaswani2017attention} without pretraining, strong pretrained AR models~\textbf{MASS}~\cite{song2019mass},~\textbf{BART}~\cite{lewis2019bart} and~\textbf{ProphetNet}~\cite{yan2020prophetnet} as our baselines. Most of the AR baseline results are collected from~\textbf{GLGE}~\cite{liu2020glge}. We collect the base version (6-layers encoder, 6-layers decoder, 768 hidden sizes) baseline models results to be compared with same model size BANG  for fair comparison.  BANG AR finetuning hyper-parameters are: learning rate 1e-4, warm up steps of 1000, Adam optimizer, the maximum input and output length of 512, and a label smoothness of 0.1. We finetune BANG on each dataset for 10 epochs and select the best model according to the performance on their dev sets. For inference, we set the beam size as 4, length penalty as 1.0 and batch size as 1 to calculate the latency.

For NAR generation, we load the BANG pretrained model and finetune it with all [MASK] inputs.
We use ~\textbf{NAT}~\cite{gu2017non}, ~\textbf{iNAT}~\cite{lee2018deterministic}, ~\textbf{CMLM}~\cite{ghazvininejad2019mask} and ~\textbf{LevT}~\cite{gu2019levenshtein} as the NAR baseline models. For these baselines, we select the outputs from the first iteration as the NAR outputs if they are semi-NAR models.
For NAR finetuning experiments, the hyper-parameters are the same as AR finetuning except the number of finetuning epochs, since NAR finetuning need more epochs to converge. We finetune BANG for 50 epochs and save a checkpoint for every 10 epochs. 
We select the best checkpoint based on the performance on dev set. For inference, we set the no-repeat-ngram hyper-parameter as 2 to merge consecutive same tokens. Since we consider the first [EOS] token as the end signal rather than predicting the target sentence length, we set the maximal output length as 50, 85, 30 for SQuAD, XSum and PersonaChat, respectively.

For semi-NAR generation, we load the BANG pretrained model and finetune it with multiple predicting streams to simultaneously support AR inference and NAR inference. We set the maximum output length and predicting stream numbers as 30, 30, 40 for SQuAD, PersonaChat, XSum, respectively. 
In other words, the model produced from multi-stream finetuning can simultaneously support AR inference, NAR inference and semi-NAR inference. 
For semi-NAR, we predict the first $n_{ar}$ tokens using a token-by-token manner sequentially, followed by predicting the last $n_{nar} = n - n_{ar}$ tokens in parallel via a single step. We set $n_{ar}$ as 5, and $n_{nar}$ as 25, 25, 35 for SQuAD, PersonaChat and XSum, respectively. The semi-NAR strategy in BANG is quite flexible to support different sequential and parallel combinations, and we leave it as future work for further exploration. For semi-NAR baselines, we choose ~\textbf{InsT}~\cite{stern2019insertion}, ~\textbf{iNAT}~\cite{lee2018deterministic}, ~\textbf{CMLM}~\cite{ghazvininejad2019mask}, ~\textbf{LevT}~\cite{gu2019levenshtein} and set the maximum iteration steps as 10 (one decoding followed by up to nine iterative refinements). 

For NAR and semi-NAR experiments, we do not utilize knowledge distillation~\cite{hinton2015distilling, kim2016sequence} from AR models. For NAR generation, knowledge distillation usually refers to finetuning with the inference outputs from an AR model~\cite{kim2016sequence, gu2017non}, in other words, the finetuning input keeps unchanged, while the targt sequence is replaced with the output from an AR model. We do not utilize knowledge distillation for two reasons: We want to use the same fine-tuning corpus as AR models to compare the performance difference; Knowledge distillation with different generated results and different AR models will result in different NAR performance and it's hard for other researchers to get the same finetuning corpus for reproducing our experiments. NAR and semi-NAR baseline models are integrated into Fairseq library~\cite{ott2019fairseq} and we use the default hyper-parameters\footnote{\href{https://github.com/pytorch/fairseq/blob/v0.9.0/examples/nonautoregressive_translation/scripts.md}{Fairseq NAR Baseline Models}} to carry out experiments.

For all downstream tasks, we use 8 NVIDIA Tesla V100 GPUs for finetuning and one single V100 GPU for inference. All the experiments are conducted on the Fairseq~\cite{ott2019fairseq} v0.9.0 codebase and we use the built-in time statistics function to calculate the per-sample inference latency.

\subsubsection{Datasets}

We conduct experiments on following three popular generation benchmarks:

\noindent\textbf{XSum} \citep{narayan2018don} contains 227K online article and single sentence summary pairs from the British Broadcasting Corporation (BBC). The average input and output lengths are 358.5 and 21.1, respectively.

\noindent\textbf{SQuAD 1.1}~\citep{rajpurkar2016squad} is a dataset created for machine reading comprehension. After preprocessing, the dataset contains 98K $\langle$answer, passage, question$\rangle$ data triples. Input is formatted as $\langle$answer [SEP] passage$\rangle$ following GLGE. The average input and output lengths are 149.4 and 11.5, respectively.

\noindent\textbf{PersonaChat}~\citep{zhang2018personalizing} is a dataset created for multi-turn conversation with personalizing profiles . After preprocessing, the dataset contains 150k $\langle$persona profile description text, conversation history, response$\rangle$ data triples. Input is formatted as $\langle$profile [SEP] conversation history$\rangle$ following GLGE. The average input and output lengths are 120.8 and 11.8, respectively.

\subsection{Main Results} \label{sec:exp2}

\begin{table*}[h]
\caption{Results of semi-NAR, NAR and AR on SQuAD 1.1.}
\small 
\begin{tabular}{llccccc}
\hline
 Pattern & Methods&  \multicolumn{4}{c}{SQuAD 1.1}&Latency \\                   
 & & ROUGE-L&BLEU-4&METEOR & OVERALL &ms/Sample  \\ \hline                                    
\multirow{5}{*}{Semi-NAR}
& InsT~\cite{stern2019insertion} &  29.98&2.34&8.15& 13.49 (+0.00) & 67.61 (4.3x)\\ 
& iNAT~\cite{lee2018deterministic}  &  32.34&3.16&9.18 & 14.89 (+1.40) & 31.59 (2.0x) \\ 
& CMLM~\cite{ghazvininejad2019mask} &  29.60&3.89&9.70 &14.40 (+0.91)& 106.84 (6.8x)\\
& LevT~\cite{gu2019levenshtein}  &  30.81&2.68&9.40&14.30 (+0.81)& 116.41 (7.4x)\\
& BANG  & \textbf{47.39}&\textbf{17.62}&\textbf{21.69} &\textbf{28.90 (+15.41)}& 111.11 (7.1x) \\ \hline
\multirow{5}{*}{NAR}
& NAT~\cite{gu2017non} &   31.51&2.46&8.86 &14.28 (+0.02)& 17.11 (1.1x) \\ 
& iNAT~\cite{lee2018deterministic} & 32.44&2.33&8.84 &14.54 (+0.28)& 16.52 (1.1x) \\ 
& CMLM~\cite{ghazvininejad2019mask}   & 31.58&2.51&8.85 &14.31 (+0.05)& 16.41 (1.0x) \\
& LevT~\cite{gu2019levenshtein}  &  31.38&2.27&9.14 &14.26 (+0.00)& 27.52 (1.8x) \\
&BANG  &   \textbf{44.07}&\textbf{12.75}&\textbf{18.99} &\textbf{25.27 (+11.01)} & \textbf{15.69 (1.0x)}  \\ \hline  
\multirow{6}{*}{AR}
& Transformer~\cite{vaswani2017attention}  &  29.43   & 4.61 &  9.86 & 14.63(+0.00)  & 159.49(10.2x) \\
& MASS~\cite{song2019mass} &   \textbf{49.48}&20.16&\textbf{24.41}  & 31.35 (+16.72) & N/A\\
& BART~\cite{lewis2019bart} &   42.55&17.08&23.19  &27.61 (+12.98)& N/A \\
& ProphetNet~\cite{yan2020prophetnet} &   48.00&19.58&23.94  &30.51 (+15.88)& N/A \\
& BANG & 49.32&\textbf{21.40}&24.25 & \textbf{31.66 (+17.03)}& N/A\\ \hline
\end{tabular}
\label{table.result.qg}
\end{table*}

\begin{table*}[h]
\caption{Results of semi-NAR, NAR and AR on XSum.}
\small 
\begin{tabular}{llccccc}
\hline
 Pattern & Methods&  \multicolumn{4}{c}{XSum}&Latency \\                   
 & & ROUGE-1&ROUGE-2&ROUGE-L&OVERALL & ms/Sample  \\ \hline                                    
\multirow{5}{*}{Semi-NAR}
& InsT~\cite{stern2019insertion} &  17.65 & 5.18 & 16.05  &12.96 (+0.00) &  63.37 (4.0x) \\ 
& iNAT~\cite{lee2018deterministic}  &  26.95 & 6.88 & 22.43 &18.75 (+5.79)&  31.27 (2.0x)  \\ 
& CMLM~\cite{ghazvininejad2019mask} &  29.12 & 7.70 & 23.04  &19.95 (+6.99)& 113.64 (7.1x) \\
& LevT~\cite{gu2019levenshtein}  &  25.33 & 7.40 & 21.48 & 18.07 (+5.11) & 101.01 (6.3x) \\
& BANG  & \textbf{34.71} & \textbf{11.71} & \textbf{29.16} & \textbf{25.19 (+12.23)} & 109.77 (6.9x) \\ \hline
\multirow{5}{*}{NAR}
& NAT~\cite{gu2017non} &   24.04 & 3.88 & 20.32 &16.08 (+0.22)& 17.47 (1.1x) \\ 
& iNAT~\cite{lee2018deterministic} & 24.02 & 3.99 & 20.36 &16.12 (+0.26)&  16.94 (1.1x) \\ 
& CMLM~\cite{ghazvininejad2019mask}   & 23.82 & 3.60 & 20.15 &15.86 (+0.00)& 16.88 (1.1x)  \\
& LevT~\cite{gu2019levenshtein}  &  24.75 &  4.18 & 20.87 & 16.60 (+0.74) & 27.72 (1.7x) \\
&BANG  &   \textbf{32.59} & \textbf{8.98} & \textbf{27.41} &\textbf{22.99 (+7.13)} & \textbf{15.97 (1.0x)}  \\ \hline
\multirow{6}{*}{AR}
& Transformer~\cite{vaswani2017attention}  &  30.66    & 10.80 & 24.48 & 21.98(+0.00) & 262.47(16.4x) \\ 
& MASS~\cite{song2019mass} &   39.70 &  17.24 & 31.91  & 29.62 (+7.64) & N/A \\
& BART~\cite{lewis2019bart} &   38.79 & 16.16 & 30.61  & 28.52 (+6.54) &  N/A\\
& ProphetNet~\cite{yan2020prophetnet} &   39.89 &  17.12 & 32.07  & 29.69 (+7.71)& N/A \\
& BANG & \textbf{41.09} & \textbf{18.37} & \textbf{33.22} & \textbf{30.89 (+8.91)} &  N/A \\  \hline
\end{tabular}
\label{table.result.XSum}
\end{table*}

\begin{table*}[h]
\caption{Results of semi-NAR, NAR, and AR on PersonaChat. D-1(Distinct-1), D-2(Distinct-2) are multiplied by 100 according to GLGE.}
\small 
\begin{tabular}{llcccccc}
\hline
 Pattern & Methods&  \multicolumn{5}{c}{PersonaChat}&Latency \\                   
& & BLEU-1&BLEU-2&D-1&D-2 & OVERALL& ms/Sample \\ \hline                                                                   

\multirow{5}{*}{Semi-NAR}
& InsT~\cite{stern2019insertion}  &  12.63&9.43&0.1&0.3 & 5.62 (+0.00) & 65.27 (4.4x) \\ 
& iNAT~\cite{lee2018deterministic} & 41.17&32.13&0.1&1.1  & 18.63 (+13.01)& 43.25 (2.9x)  \\ 
& CMLM~\cite{ghazvininejad2019mask} &   \textbf{44.38}&\textbf{35.18}&0.1&0.8 & 20.12 (+14.50) &105.82 (7.1x)\\ 
& LevT~\cite{gu2019levenshtein} &  24.89&18.94&0.1&0.6 &11.13 (+5.51)& 80.26 (5.4x) \\ 
& BANG  & 39.82 & 30.72  & \textbf{1.9} & \textbf{14.2}  & \textbf{21.66 (+16.04)} & 109.17 (7.3x) \\  \hline
\multirow{5}{*}{NAR}
& NAT~\cite{gu2017non} &  31.53 & 24.17 & 0.1& 0.8 & 14.15 (+2.20)& 17.86 (1.2x) \\ 
& iNAT~\cite{lee2018deterministic} & 30.56&23.38&0.1&0.7 & 13.69 (+1.74)& 16.40 (1.1x) \\ 
& CMLM~\cite{ghazvininejad2019mask} &\textbf{31.44}&\textbf{24.06}&0.1&0.6 &14.05 (+2.10)& 16.26 (1.1x) \\
& LevT~\cite{gu2019levenshtein} & 26.92 & 20.47 & 0.0 & 0.4 &11.95 (+0.00)& 27.56 (1.9x)\\ 
& BANG & 31.11&23.90&\textbf{2.5}&\textbf{22.7} & \textbf{20.05 (+8.10)} &\textbf{14.89 (1.0x)} \\ \hline
\multirow{6}{*}{AR}
&Transformer~\cite{vaswani2017attention} & 41.56  & 32.95 & 0.3 & 0.8 & 18.90(+0.00) & 138.31(9.3x)  \\ 
&MASS~\cite{song2019mass} & 41.06&35.75&1.4&6.9&21.28 (+2.38)& N/A \\
&BART~\cite{lewis2019bart} &  \textbf{47.60}&\textbf{39.36}&1.1&6.1&\textbf{23.54(+4.64)}& N/A \\
& ProphetNet~\cite{yan2020prophetnet} &   46.00 &  38.40 & 1.3  & 7.3 & 23.25 (+4.35)& N/A \\
&BANG & 45.77 & 35.54 &\textbf{1.4}&\textbf{8.4} & 22.78 (+3.88) & N/A \\  \hline
\end{tabular}
\label{table.result.dialog}
\end{table*}

We present the results for question generation task in Table~\ref{table.result.qg}, summarization task in Table~\ref{table.result.XSum}, and dialog task in Table~\ref{table.result.dialog}. BANG achieves significant performance improvements on all tasks consistently in both the NAR and the semi-NAR settings. Compared with the best semi-NAR baselines, BANG achieves absolute improvements of 14.01 and 5.24 in overall scores of SQuAD and XSum, respectively. In addition, BANG achieves absolute improvements of 10.73, 6.39 and 5.90 in overall scores of SQuAD, XSUM and PersonaChat compared with the best NAR baselines, respectively. 
This clearly demonstrates the effectiveness of the BANG pretraining. The NAR or semi-NAR results based on BANG are  comparable or even better than the AR generation methods without pretraining.

From Table~\ref{table.result.qg}, we can see that via BANG pretraining, the results on semi-NAR and NAR generation are significantly improved. 
Meanwhile, the results on AR generation via BANG pretraining are comparable with strong pretrained baselines. On the other hand, both BANG semi-NAR and NAR models outperforms Transformer AR baselines with obviously reduced latency. For example, the per sample inference latency of Transformer model is 159.49ms while that of  BANG NAR is only 15.69ms , which amounts to a 10.2 times speedup. In terms of the inference speed, we can see that BANG is slightly better than other NAR generation models. This speedup is mainly due to the fact that BANG NAR uses the special token [SEP] as the ending mark, without the need to predict the length.

Similar to the Question Generation tasks, the improvement on the XSum in Table~\ref{table.result.XSum} is also significant. In all the three categories semi-NAR, NAR and AR, BANG can outperform all of the baselines consistently. Meanwhile, BANG's NAR and semi-NAR results are better than AR generation models without pretraining. Furthermore, we observe an interesting result when studying both the SQuAD and XSum together: BANG semi-NAR consistently outperforms NAR in both tasks, while the baseline models show that semi-NAR via iterative refinements can help XSum summarization but often hurt SQuAD question generation.

For the results on PersonaChat in Table~\ref{table.result.dialog}, we focus on the dialog outputs diversity Distinct-1 and Distinct-2 metrics. For dialog generation, the outputs are very open such that the outputs diversity is deeply concerned to avoid boring, or useless dialog responses. We can see the D-1 and D-2 of models without pretraining are very low, which means even the inputs are different, their outputs are highly overlapped or just repetitive. For NAR baseline models, we observe the outputs are often composed of ``i, a, an, the" and punctuation, which may share some repetitive parts with target sequence to achieve a reasonable B-1 or B-2 score, but lacking meaningful responses. For semi-NAR baseline models, we observe the iterative refinement can help produce a fluent response. However, the generated responses are often unrelated to the dialog context with a low accuracy. Baseline AR models have high B-1 and B-2 scores via producing some common phrases like ``i have", ``have a", ``a lot", but meaningless dialog responses. On contrary, from BANG, we observe meaningful response and high diversity in the response, which makes it achieve a good diversity score consistently in all three categories: the high-speed NAR generation, the high-performance AR generation, and the semi-NAR generation. We list more examples in the Appendix for a detailed illustration.

\subsection{NAR Results with AR, NAR, BANG Pretraining} \label{sec.abalation.pretrain}
Note that NAR baseline models in~\S~\ref{sec:exp2} are not pretrained. Here we provide AR and NAR pretrained models to compare with BANG. We load the parameters from MASS as the AR pretrained model and carry out NAR finetuning. NAR pretraining setting is same as BANG pretraining, except that we replace the cross-stream visible multi-stream decoder with a single decoder filled with [MASK] to make it consistent with the NAR finetuning. 

\begin{table}[h]
\caption{NAR results on SQuAD 1.1 question generation with different pretraining strategy.}
\small
\begin{tabular}{lccc}
\hline
 Pretrain Strategy             & ROUGE-L & BLEU-4 & METEOR \\ \hline
No-pretrain     & 30.69   & 2.07   & 8.22   \\ 
AR(MASS) & 40.20 & 9.50 & 16.00 \\
NAR    & 41.15   & 11.38  & 17.63  \\ 
BANG & 44.07   & 12.75  & 18.99  \\ \hline
\end{tabular}
\label{table.result.pretrain.qg}
\end{table}

\begin{table}[h]
\caption{NAR results on XSum summarization with different pretraining strategy.}
\small
\begin{tabular}{lccc}
\hline
 Pretrain Strategy          & ROUGE-1 & ROUGE-2 & ROUGE-L \\ \hline
No-Pretrain     & 23.86   & 4.17    & 20.12   \\ 
AR(MASS) & 26.10 & 6.30 & 22.50 \\
NAR    & 30.43   & 7.66    & 25.50    \\ 
BANG & 32.50    & 8.92    & 27.35   \\ \hline
\end{tabular}
\label{table.result.pretrain.XSum}
\end{table}

The results for NAR finetuning results with different pretraining settings are shown in Table~\ref{table.result.pretrain.qg} and Table~\ref{table.result.pretrain.XSum}. First, pretraining improves NAR finetuning results significantly and consistently in both tasks. Second, AR pretrained models perform worse than NAR pretrained models because AR pretraining has inconsistent language model between pretraining and finetuning. Third, BANG, with the same NAR finetuning procedure and model architecture(the n-stream archtecture difference only exists in the pretraining, and in the NAR finetuning procedure, two models have the exactly same architecture), consistently outperforms the NAR pretrained model. This clearly demonstrates that the proposed pretraining strategy via bridging both AR and NAR is critical to achieve a better performance in NAR generation. Lastly, the improvements of BANG are consistent and significant in both tasks across all metrics.

\section{Related Work}

AR models have been developed for a long time. Recent works show that pretraining on large scale text will guarantee a consistent improvements on downstream generation tasks. GPT series work~\cite{radford2018improving, radford2019language, brown2020language} pretrains decoders with the task of next token prediction and convert different downstream tasks into language models. MASS~\cite{song2019mass} masks continuous words' spans from input sentences to predict. BART~\cite{lewis2019bart} uses denoising task to pretrain Transformer. ProphetNet~\cite{yan2020prophetnet} deploys future tokens' prediction to enhance AR generation ability. DialogGPT~\cite{zhang2019dialogpt} is pretrained for conversational response generation. XLNet~\cite{yang2019xlnet} utilizes AR pretraining for downstream natural language understanding tasks, and we also borrow 2-stream strategy from XLNet for BANG cross-stream visible n-stream decoder.

NAR and semi-NAR models are proposed to accelerate natural language generation. NAT~\cite{gu2017non} is proposed as a NAR generation model to decode the whole target sequence at one time step. iNAT~\cite{lee2018deterministic} refines outputs with multi-turn post-processing.
InsT~\cite{stern2019insertion} predicts inserting positions and inserting tokens at each iteration. CMLM~\cite{ghazvininejad2019mask} firstly predict all target words with NAR generation and maskout-regenerate low confidence words. LevT~\cite{gu2019levenshtein} considers two basic operations insertion and deletion at each iteration. 

Some works propose to use AR generation facilitate NAR generation and some work propose future tokens' prediction to facilitate AR generation. Curriculum learning~\cite{bengio2009curriculum} is used to benefit NAR generation from AR generation such as~\citet{guo2020fine}. Future tokens' prediction is used to benefit AR generation such as~\citet{yan2020prophetnet, goodman2020teaforn}. Recently, algorithms~\cite{tian2020train, mansimov2019generalized} to support different generation patterns with the same model start to be researched.

BANG benefits from these different NAR and AR generation models to support NAR, AR, semi-NAR generation.

\section{Conclusion}
We propose a new natural language generation pretraining model named BANG. BANG bridges NAR and AR generation with cross-stream visible n-stream strategy and large scale pretraining. BANG supports NAR, AR and semi-NAR generation with the same pretrained model. Experiments show that BANG can significant improve the NAR and semi-NAR generation performance, and provide comparable results with strong AR pretraining models. NAR and semi-NAR generation can be applied to general natural language generation tasks with acceptable performance with BANG pretraining. BANG is powerful and flexible to support more diverse semi-NAR generation strategies and finetuning strategies, which we leave as future work.

\newpage

\bibliography{main}
\bibliographystyle{icml2021}

\end{document}


\begin{table*}[h]
    \small
    \centering
    \begin{tabularx}{\textwidth}{lX}
        \toprule
        \multicolumn{2}{c}{\textbf{SQuAD 1.1}} \\    \bottomrule
        \textbf{Input}    &   Forbes {[}SEP{]} A self - described `` modern - day feminist " , Beyoncé creates songs that are often characterized by themes of love , relationships , and monogamy , as well as female sexuality and empowerment. ......  Forbes magazine also listed her as the most powerful female musician of 2015 .    \\    \bottomrule
        \textbf{Golden}    &   which magazine declared her the most dominant woman musician ?
   \\    \bottomrule
        \textbf{NAT}    &   where is the music of s in called ?
   \\    \bottomrule
        \textbf{CMLM NAR}    &   what is the is the music in music ?
   \\    \bottomrule
        \textbf{CMLM semi-NAR}    &   what is the name of the most popular music ?
   \\    \bottomrule
        \textbf{BANG NAR}    &   who magazine listed her as the most powerful musician ?
   \\    \bottomrule
        \textbf{BANG semi-NAR}    &   who listed her as the most powerful female musician of 2015 ?
   \\    \bottomrule
        \textbf{BANG AR}    &   which magazine listed beyonce as the most powerful female musician in 2015 ?
   \\    \bottomrule
           \multicolumn{2}{c}{\textbf{XSum}} \\    \bottomrule
        \textbf{Input}    &   She became Kenya's first high-profile athlete to fail a test, when she tested positive for performance-enhancing drugs in September.Jeptoo, 33, says she may have been prescribed some banned substances at a local hospital after a road accident.She has become the 45th Kenyan athlete to have failed a doping test. ...... She has won the previous three Boston and two Chicago marathons and also previously won the Stockholm, Paris, Milan and Lisbon marathons.
    \\    \bottomrule
        \textbf{Golden}    &   kenya's rita jeptoo, winner of the boston and chicago marathons, has been banned for two years after failing a drugs test.
   \\    \bottomrule
        \textbf{NAT}    &   kenya kenyan s kenya - kenya has , the a doping .
  \\    \bottomrule
        \textbf{CMLM NAR}    &   kenyan s je -oo has banned , the - doping .
   \\    \bottomrule
        \textbf{CMLM semi-NAR}    &   kenyan olympic gold medallist laura jefioo has been banned from the country of an anti - doping .
   \\    \bottomrule
        \textbf{BANG NAR}    &   kenyan marathon runner rita jeptoo has been given for two year ban after failing doping athletics .
   \\    \bottomrule
        \textbf{BANG semi-NAR}    &   kenyan marathon runner kiba jeptoo has been banned for two years for failing doping .
   \\    \bottomrule
        \textbf{BANG AR}    &   kenya ' s world marathon champion lydia jeptoo has been banned for two years by athletics kenya for failing a doping test .
   \\    \bottomrule
    \end{tabularx}
    \caption{Case study for different NAR, AR and semi-NAR models on SQuAD 1.1 and XSum.}\label{table.casestudy}
\end{table*}
\section{Case Study}\label{sec.casestudy}
In this section, we choose two samples from SQuAD question generation and XSum summarization to help illustrate how BANG helps NAR and semi-NAR generation in Table~\ref{table.casestudy}. We also provide more examples for PersonaChat in Table~\ref{table.ablation.PersonaChat}. We choose NAT and CMLM as NAR and semi-NAR baselines,  respectively, since NAT is the first NAR generation model and CMLM has the best performance according to the experimental results in paper results.

For results on SQuAD in Table~\ref{table.casestudy}, we can see that SQuAD question generation is easier than XSum with more fluent NAR generation because of the shorter output length and the question sentence format. Baseline NAR output sentence structures seem reasonable but the described items are wrong. We remark that with iterative refinements the results become more fluent, but the modification is often only based on the first generated outputs. For example, the example continues to describe the wrong item ``music" rather than ``musician". Besides, although CMLM semi-NAR outputs are fluent with refinements, the ability of understanding the task target is still weak and heavily relies on the output in the first iteration. On the contrary, BANG is able to raise proper question on correct object. Although BANG NAR generation mixes ``who listed" and ``which magazine listed" into "who magazine listed" because of the weakness of NAR generation ability, BANG semi-NAR generation fixes that problem into ``who listed her as the" to serve as a high-quality sub-sequence hint. In the AR generation result, BANG even finds ``her" represents ``beyonce" from the input context.

For results on XSum in Table~\ref{table.casestudy}, we can see that baseline NAR models nearly fail in generating meaningful sentences. Baseline NAR results are composed with key phrases and key words while BANG NAR generation are fluent with insignificant errors. CMLM semi-NAR results are fluent and close to target sequence via iterative modification. BANG semi-NAR output has a mistake at the end of the sentence where ``failing doping" should be ``failing a doping test". BANG AR output contains nearly all the important details. A common problem for all the outputs is about the runner name ``jeptoo". In XSum training samples, they often come with a complete names composed with the first name and the second name. Models learn to generate complete human names. In this given input, only the second name ``jeptoo" is given and the models have no idea about her first name. All the models fake a first name for her to generate a complete description, which shows that natural language generation models have a bias on the training data and may fabricate some details with maximum likelihood.

For results on PersonaChat in Table~\ref{table.ablation.PersonaChat}, we can see though the baseline models have high B-1 B-2 evaluation results in Table~\ref{table.result.dialog}, they fail to generate meaningful dialog responses. NAR baseline outputs are composed with common words while semi-NAR baseline outputs are composed with common sentences. Although the common words, phrases, sentences have n-gram overlap with target to have high B-1 and B-2 scores, they are hard to compose meaningful responses. On the contrary, no matter what the generation pattern is, BANG models seek to generate diverse and meaningful responses. 

\begin{table*}[h]
    \small
    \centering
    \begin{tabularx}{\textwidth}{lX}
        \toprule
   \multirow{8}{*}{Input} 
                       &  i have two dogs and one cat . ...{[}SEP{]} ...  do you have pets ? no i do not , do you ?          \\
                       &  i have two dogs and one cat . ... {[}SEP{]} ... nice , where do you live ? i resign in north dakota                 \\
                       &   i like to make crafts . ... {[}SEP{]} ...  i live in a small town in ohio .           \\
                       &  i work in a factory .... {[}SEP{]} ... i live in a small town in ohio . so we are semi close neighbors . \\
                       &  i like to dance at the club ....{[}SEP{]} hi there , how are you today ?                          \\
                       &  i just had surgery . ...{[}SEP{]} ...  i am great ! just got home from working with dogs all day !          \\
                       &  i like to dance at the club . ... {[}SEP{]} ...  oh you have dogs ? i have two chow chow .                \\
                       &  i like to dance at the club .... {[}SEP{]} ...  i just had a surgery not long ago .
    \\    \bottomrule   
    \multirow{8}{*}{Golden} 
                       &   yes . two dogs and a cat . they are my babies .      \\
                       &  i live in texas . i love riding my bike here .      \\
                       &  so we are semi close neighbors .   \\
                       &   seems like it . have you been to the rock hall ?  \\
                       &  i am great ! just got home from working with dogs all day !                      \\
                       &  oh you have dogs ? i have two chow chow .        \\
                       &  i love dogs ! i have dogs and i get to train them at work , too !       \\
                       &  i hope you feel better soon . what do you like to do for fun ?
     \\    \bottomrule  
     \multirow{8}{*}{NAT} 
                       & i is , the ?     \\
                       &  yes , i a cat     \\
                       &  that is . do you any pets ?    \\
                       &  that is . do you any pets ? \\
                       & i am good , how are you ?                       \\
                       & i am i . just . i . my dogs .        \\
                       & i , i my you .       \\
                       &  i is i a dogs .
     \\    \bottomrule  
     \multirow{8}{*}{CMLM NAR} 
                       & i , you a you ?       \\
                       & i , i . i you a you .       \\
                       & i is i . i . i the you .    \\
                       & i is i . i the live .  \\
                       & i am doing , how are you ?                      \\
                       & i am i . you ?        \\
                       & i ! . i do a you .       \\
                       &  i ! . i . i a dogs .
     \\    \bottomrule  
     \multirow{8}{*}{CMLM semi-NAR} 
                       & i have a few years , i have a lot of my dog .       \\
                       & i have a few dogs , i have a lot of them .       \\
                       & i have a lot of dogs . i have a lot of them .    \\
                       &  i have a lot of dogs . i have a lot of them .  \\
                       & i am good , how are you ?                      \\
                       & i am doing well , i am doing well . how are you ?        \\
                       & i am a teacher . i am a big teacher . but i for a lot .       \\
                       &  i am a teacher . i just am a lot job for work a lot .
     \\    \bottomrule  
     \multirow{8}{*}{BANG NAR} 
                       &  yes . i have and a cat . is      \\
                       &  i live in north dakota . i ride bike .       \\
                       &  maybe you like ?   \\
                       &  i is close . you about ? .  \\
                       & i am doing great thanks !                      \\
                       & oh nice ! i have dogs have big dogs .        \\
                       & yes ! dogs ! my sweet !       \\
                       &  surgery , surgery that sounds bad ? 
      \\    \bottomrule  
     \multirow{8}{*}{BANG semi-NAR} 
                       &  yes , i have two dogs and a cat .      \\
                       &  i live in north carolina . i ride bike .      \\
                       &  i have never been there . i homes there   \\
                       &  that is good . do you have any hobbies ? \\
                       & i am good . just got back from .  \\
                       & oh , i love dogs ! i have two dogs .        \\
                       & i have two ! i have a sweet tooth !       \\
                       &  i am so sorry to hear that . what you do ?   
      \\    \bottomrule                             
     \multirow{8}{*}{BANG AR} 
                       & yes , i have two dogs and a cat .       \\
                       & i live in california . i love riding my bike there .       \\
                       & do you have any hobbies ? i like to make crafts .    \\
                       & yes , we are . what do you do for a living ?  \\
                       & i am great ! just got back from dance practice . you ?                       \\
                       & that sounds like a lot of fun . i have two dogs .        \\
                       & yes , i do ! i have a big sweet tooth . what kind of food do you like ?       \\
                       &  oh , i am sorry to hear that . what kind of surgery ?
    \\    \bottomrule  
    \end{tabularx}
    \caption{Case study for different NAR, AR and semi-NAR models on PersonaChat.}\label{table.ablation.PersonaChat}
\end{table*}


\begin{table*}[h]
\small 
\begin{tabular}{llccccc}
\hline
 Pattern & Methods&  \multicolumn{4}{c}{SQuAD 1.1}&Latency \\                   
 & & ROUGE-L&BLEU-4&METEOR & OVERALL &ms/Sample  \\ \hline                                    
\multirow{5}{*}{Semi-NAR}
& InsT~\cite{stern2019insertion} &  29.98&2.34&8.15& 13.49 (+0.00) & 67.61 (4.3x)\\ 
& iNAT~\cite{lee2018deterministic}  &  32.34&3.16&9.18 & 14.89 (+1.40) & 31.59 (2.0x) \\ 
& CMLM~\cite{ghazvininejad2019mask} &  29.60&3.89&9.70 &14.40 (+0.91)& 106.84 (6.8x)\\
& LevT~\cite{gu2019levenshtein}  &  30.81&2.68&9.40&14.30 (+0.81)& 116.41 (7.4x)\\
& BANG$_{base}$  & \textbf{47.39}&\textbf{17.62}&\textbf{21.69} &\textbf{28.90 (+15.41)}& 111.11 (7.1x) \\ 
& BANG$_{large}$  & 48.69 & 18.75 & 22.27 & 29.90 & 207.04 \\ \hline
\multirow{5}{*}{NAR}
& NAT~\cite{gu2017non} &   31.51&2.46&8.86 &14.28 (+0.02)& 17.11 (1.1x) \\ 
& iNAT~\cite{lee2018deterministic} & 32.44&2.33&8.84 &14.54 (+0.28)& 16.52 (1.1x) \\ 
& CMLM~\cite{ghazvininejad2019mask}   & 31.58&2.51&8.85 &14.31 (+0.05)& 16.41 (1.0x) \\
& LevT~\cite{gu2019levenshtein}  &  31.38&2.27&9.14 &14.26 (+0.00)& 27.52 (1.8x) \\
&BANG$_{base}$  &   \textbf{44.07}&\textbf{12.75}&\textbf{18.99} &\textbf{25.27 (+11.01)} & \textbf{15.69 (1.0x)} 
\\  
& BANG$_{large}$ & 45.13 & 14.02 & 19.84 & 26.33 & 29.84 \\ \hline
\multirow{6}{*}{AR}
& Transformer$_{base}$~\cite{vaswani2017attention}  &  29.43   & 4.61 &  9.86 & 14.63  & 159.49 \\
& MASS$_{base}$~\cite{song2019mass} &   \textbf{49.48}&20.16&\textbf{24.41}  & 31.35 (+15.86) & N/A\\
& BART$_{base}$~\cite{lewis2019bart} &   42.55&17.08&23.19  &27.61 (+12.12)& N/A \\
& ProphetNet$_{base}$~\cite{yan2020prophetnet} &   48.00&19.58&23.94  &30.51 (+15.02)& N/A \\
& BANG$_{base}$ & 49.32&\textbf{21.40}&24.25 & \textbf{31.66 (+16.17)}& N/A\\
& Transformer$_{large}$~\cite{vaswani2017attention}  &     30.73&4.80&10.93 & 15.49 (+0.00) & 233.10 (14.9x)\\
& MASS$_{middle}$~\cite{song2019mass} & 49.9 & 21.3  & 25.2 & 32.13 &  N/A \\
& BART$_{large}$~\cite{lewis2019bart} & 50.3 & 22.0 & 26.4 & 32.9  &  N/A  \\
& ProphetNet$_{large}$~\cite{yan2020prophetnet} & 51.5 & 22.5 & 26.4 & 33.47 &   N/A \\
& BANG$_{large}$ & 51.31 & 23.41 & 25.84 & 33.52 & N/A \\ \hline
\end{tabular}
\caption{Results of semi-NAR, NAR and AR on SQuAD 1.1.}
\label{table.result.qg}
\end{table*}

\begin{table*}[h]
\small 
\begin{tabular}{llccccc}
\hline
 Pattern & Methods&  \multicolumn{4}{c}{XSum}&Latency \\                   
 & & ROUGE-1&ROUGE-2&ROUGE-L&OVERALL & ms/Sample  \\ \hline                                    
\multirow{5}{*}{Semi-NAR}
& InsT~\cite{stern2019insertion} &  17.65 & 5.18 & 16.05  &12.96 (+0.00) &  63.37 (4.0x) \\ 
& iNAT~\cite{lee2018deterministic}  &  26.95 & 6.88 & 22.43 &18.75 (+5.79)&  31.27 (2.0x)  \\ 
& CMLM~\cite{ghazvininejad2019mask} &  29.12 & 7.70 & 23.04  &19.95 (+6.99)& 113.64 (7.1x) \\
& LevT~\cite{gu2019levenshtein}  &  25.33 & 7.40 & 21.48 & 18.07 (+5.11) & 101.01 (6.3x) \\
& BANG$_{base}$  & \textbf{34.71} & \textbf{11.71} & \textbf{29.16} & \textbf{25.19 (+12.23)} & 109.77 (6.9x) \\
& BANG$_{large}$  &  37.69  & 14.04 & 31.90 & 27.88 & 212.31 \\ \hline
\multirow{5}{*}{NAR}
& NAT~\cite{gu2017non} &   24.04 & 3.88 & 20.32 &16.08 (+0.22)& 17.47 (1.1x) \\ 
& iNAT~\cite{lee2018deterministic} & 24.02 & 3.99 & 20.36 &16.12 (+0.26)&  16.94 (1.1x) \\ 
& CMLM~\cite{ghazvininejad2019mask}   & 23.82 & 3.60 & 20.15 &15.86 (+0.00)& 16.88 (1.1x)  \\
& LevT~\cite{gu2019levenshtein}  &  24.75 &  4.18 & 20.87 & 16.60 (+0.74) & 27.72 (1.7x) \\
&BANG$_{base}$  &   \textbf{32.59} & \textbf{8.98} & \textbf{27.41} &\textbf{22.99 (+7.13)} & \textbf{15.97 (1.0x)}  \\
&BANG$_{large}$  &  35.69  & 11.10 & 29.83 & 25.54 &  32.95  \\ \hline
\multirow{6}{*}{AR}
& Transformer$_{base}$~\cite{vaswani2017attention}  &  30.66    & 10.80 & 24.48 & 21.98 & 262.47 \\ 
& MASS$_{base}$~\cite{song2019mass} &   39.70 &  17.24 & 31.91  & 29.62 (+7.86) & N/A \\
& BART$_{base}$~\cite{lewis2019bart} &   38.79 & 16.16 & 30.61  & 28.52 (+6.76) &  N/A\\
& ProphetNet$_{base}$~\cite{yan2020prophetnet} &   39.89 &  17.12 & 32.07  & 29.69 (+7.93)& N/A \\
& BANG$_{base}$ & \textbf{41.09} & \textbf{18.37} & \textbf{33.22} & \textbf{30.89 (+9.13)} &  N/A \\  
& Transformer$_{large}$~\cite{vaswani2017attention}  &     30.57 & 10.47 & 24.22  & 21.76 (+0.00) & 364.96 (22.9x)\\ 
&MASS$_{middle}$~\cite{song2019mass}  &  39.1  & 16.5 & 31.4& 29 &  N/A  \\ 
&BART$_{large}$~\cite{lewis2019bart}  &   45.1 & 22.2 & 37.2 & 34.83  &  N/A  \\ 
&ProphetNet$_{large}$~\cite{yan2020prophetnet}  &44.4    & 21.3 & 36.4& 34.03 &  N/A  \\ 
&BANG$_{large}TODO$  & 44.02 & 21.04 & 36.01  & 33.69 &  N/A  \\ \hline
\end{tabular}
\caption{Results of semi-NAR, NAR and AR on XSum.}
\label{table
.result.XSum}
\end{table*}

\begin{table*}[h]
\small 
\begin{tabular}{llcccccc}
\hline
 Pattern & Methods&  \multicolumn{5}{c}{PersonaChat}&Latency \\                   
& & B-1&B-2&D-1&D-2 & OVERALL& ms/Sample \\ \hline                                                                   

\multirow{5}{*}{Semi-NAR}
& InsT~\cite{stern2019insertion}  &  12.63&9.43&0.1&0.3 & 5.62 (+0.00) & 65.27 (4.4x) \\ 
& iNAT~\cite{lee2018deterministic} & 41.17&32.13&0.1&1.1  & 18.63 (+13.01)& 43.25 (2.9x)  \\ 
& CMLM~\cite{ghazvininejad2019mask} &   \textbf{44.38}&\textbf{35.18}&0.1&0.8 & 20.12 (+14.50) &105.82 (7.1x)\\ 
& LevT~\cite{gu2019levenshtein} &  24.89&18.94&0.1&0.6 &11.13 (+5.51)& 80.26 (5.4x) \\ 
& BANG$_{base}$  & 39.82 & 30.72  & \textbf{1.9} & \textbf{14.2}  & \textbf{21.66 (+16.04)} & 109.17 (7.3x) \\  
&BANG$_{large}$ & 39.17 & 30.15  & 2.1 &  14.4  &  21.44  & 207.90 \\ \hline
\multirow{5}{*}{NAR}
& NAT~\cite{gu2017non} &  31.53 & 24.17 & 0.1& 0.8 & 14.15 (+2.20)& 17.86 (1.2x) \\ 
& iNAT~\cite{lee2018deterministic} & 30.56&23.38&0.1&0.7 & 13.69 (+1.74)& 16.40 (1.1x) \\ 
& CMLM~\cite{ghazvininejad2019mask} &\textbf{31.44}&\textbf{24.06}&0.1&0.6 &14.05 (+2.10)& 16.26 (1.1x) \\
& LevT~\cite{gu2019levenshtein} & 26.92 & 20.47 & 0.0 & 0.4 &11.95 (+0.00)& 27.56 (1.9x)\\ 
& BANG$_{base}$ & 31.11&23.90&\textbf{2.5}&\textbf{22.7} & \textbf{20.05 (+8.10)} &\textbf{14.89 (1.0x)} \\ 
&BANG$_{large}$ & 32.58  &  25.16 &  2.5 &  24.8  &    21.26 &  29.84 \\  \hline
\multirow{6}{*}{AR}
&Transformer$_{base}$~\cite{vaswani2017attention} & 41.56  & 32.95 & 0.3 & 0.8 & 18.90 & 138.31  \\ 
&MASS$_{base}$~\cite{song2019mass} & 41.06&35.75&1.4&6.9&21.28 (+3.07)& N/A \\
&BART$_{base}$~\cite{lewis2019bart} &  \textbf{47.60}&\textbf{39.36}&1.1&6.1&\textbf{23.54(+5.33)}& N/A \\
& ProphetNet$_{base}$~\cite{yan2020prophetnet} &   46.00 &  38.40 & 1.3  & 7.3 & 23.25 (+5.04)& N/A \\
&BANG$_{base}$ & 45.77 & 35.54 &\textbf{1.4}&\textbf{8.4} & 22.78 (+4.57) & N/A \\ 
&Transformer$_{large}$~\cite{vaswani2017attention} & 38.34&33.60&0.2&0.7&18.21 (+0.00)& 204.91 (13.8x)\\ 
&MASS$_{middle}$~\cite{song2019mass} & 41.0 & 35.7 & 1.4 & 6.9  &  21.25  & N/A \\
&BART$_{large}$~\cite{lewis2019bart} & 49.9 & 40.0 & 1.3 &  8.0 & 24.8 & N/A \\
&ProphetNet$_{large}$~\cite{yan2020prophetnet} & 46.7 & 39.0 & 1.3 & 7.5  &  23.63  & N/A \\
&BANG$_{large}$ & 46.41 & 36.18 & 1.8 &  11.6 &  24.00   & N/A \\ 
\hline
\end{tabular}
\caption{Results of semi-NAR, NAR and AR on Dialog. B-1: BLEU-1, B-2: BLEU-2, D-1: Distinct-1. D-2: Distinct-2. D-1 and D-2 are multiplied by 100 according to GLGE. }
\label{table.result.dialog}
\end{table*}

\newpage
\bibliography{main}
\bibliographystyle{icml2021}